\documentclass[10pt,twocolumn,letterpaper]{article}

\usepackage{cvpr}
\usepackage{times}
\usepackage{epsfig}
\usepackage{graphicx}
\usepackage{amsmath}
\usepackage{amssymb} 
  
\newcommand\ddfrac[2]{\frac{\displaystyle #1}{\displaystyle #2}}

\usepackage[pagebackref=true,breaklinks=true,letterpaper=true,colorlinks,bookmarks=false]{hyperref}
 
\cvprfinalcopy  

\begin{document}

%%%%%%%%% TITLE
\title{EV-SegNet: Semantic Segmentation for Event-based Cameras}

\author{I\~{n}igo~Alonso\\
Universidad de Zaragoza, Spain\\
{\tt\small inigo@unizar.es}
\and
Ana C. Murillo\\
Universidad de Zaragoza, Spain\\
{\tt\small acm@unizar.es}}

\maketitle

%%%%%%%%% ABSTRACT
\begin{abstract}
Event cameras, or Dynamic Vision Sensor (DVS), are very promising sensors which have shown several advantages over frame based cameras. However, most  recent work on real applications of these cameras is focused on 3D reconstruction and 6-DOF camera tracking.
Deep learning based approaches, which are leading the state-of-the-art in visual recognition tasks, could potentially take advantage of the benefits of DVS, but some adaptations are needed still needed in order to effectively work on these cameras. 
This work introduces a first baseline for semantic segmentation with this kind of data. We build a semantic segmentation CNN based on state-of-the-art techniques which takes event information as the only input. Besides, we propose a novel representation for DVS data that outperforms previously used event representations for related tasks. 
Since there is no existing labeled dataset for this task, we propose how to automatically generate approximated semantic segmentation labels for some sequences of the DDD17 dataset, which we publish together with the model, and demonstrate they are valid to train a model for DVS data only. 
We compare our results on semantic segmentation from DVS data with results using corresponding grayscale images, demonstrating how they are complementary and worth combining.
\end{abstract}

\section{Introduction}
\label{sec:introduction}
%event cameras intro
Event cameras, or Dynamic Vision Sensor (DVS) \cite{lichtsteiner2008128},  are promising sensors which register intensity changes of the captured environment. In contrast to conventional cameras, this sensor does not acquire images at a fixed frame-rate. These cameras, as their name suggest, capture events and record a stream of asynchronous events. 
An event indicates an intensity change at a specific moment and at a particular pixel (more details on how events are acquired in Section~\ref{sec:event-representation}). Event cameras offer multiple advantages over more conventional cameras, mainly: 1) very high temporal resolution, which allows the capture of multiple events in microseconds; 2) very high dynamic range, which allows the information capture at difficult lighting environments, such as night or very bright scenarios; 3) low power and bandwidth requirements.
As Maqueda et al. \cite{maqueda2018event} emphasize, event cameras are natural motion detectors and automatically filter out any
temporally-redundant information. Besides, they show that these cameras provide richer information than just subtracting consecutive conventional images. 

\begin{figure}[!tb]
\centering
    \begin{tabular}{@{}c@{\hspace{1mm}}c@{\hspace{1mm}}c}
    \includegraphics[width=.32\linewidth]{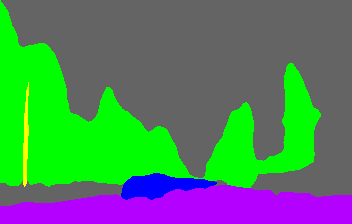} &
    \includegraphics[width=.32\linewidth]{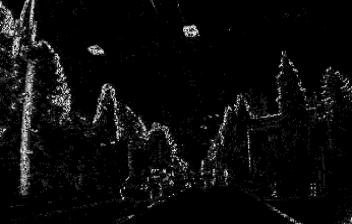} &
    \includegraphics[width=.32\linewidth]{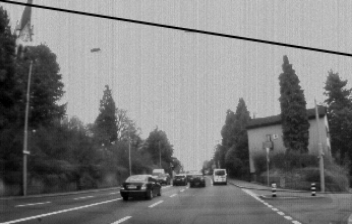}\\

    \includegraphics[width=.32\linewidth]{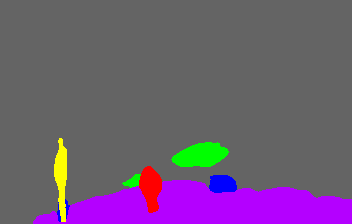} &
    \includegraphics[width=.32\linewidth]{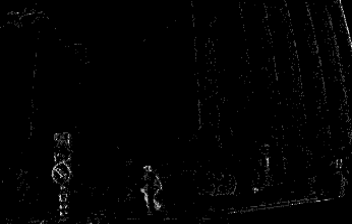} &
    \includegraphics[width=.32\linewidth]{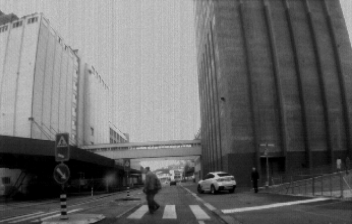} \\
    \end{tabular}
  \caption{Two examples of semantic segmentation (left) from event based camera data (middle). The semantic segmentation is the prediction of our CNN, fed only with event data. Grayscale images (right) are displayed only to facilitate visualization. Best viewed in color.}
\label{fig:introduction-results}
\end{figure}

These cameras offer a wide range of new possibilities and features that could boost solutions for many computer vision applications. 
However, new algorithms still have to be developed in order to fully exploit their capabilities, specially regarding recognition tasks. 
Most of the latest achievements based on deep learning solutions for image data, have not yet been even attempted on event cameras. 
One of the main reasons is the output of these cameras: they do not provide standard images, and %for example 
there is not yet a clearly adopted way of representing the stream of events to feed a CNN. Another challenge %reason
is the lack of labeled training data, which is key to  training most recognition models. Our work includes simple but effective novel ideas to deal with these two challenges. They could be helpful in many DVS applications, but we focus on an application not explored yet with this sensor, semantic segmentation.

This work proposes to combine the potential of event cameras with deep learning techniques on the challenging task of semantic segmentation. Semantic segmentation may intuitively seem a task much better suited to models using appearance information (from common RGB images). However, we show how, with an appropriate model and representation, event cameras provide very promising results for this task.
Figure \ref{fig:introduction-results} shows two visual results as an example of the output of our work. Our main contributions are:
\begin{itemize}
    \item First results, up to our knowledge, on semantic segmentation using DVS data. We build an Xception-based CNN that takes this data as input. Since there is no benchmark available for this problem, we propose how to generate approximated semantic segmentation labels for some sequences of the DDD17 event-based dataset. Model and data are being released.
    \item A comparison of different DVS data representation performance on semantic segmentation (including a new proposed representation that is shown to outperform existing ones), and an analysis of benefits and drawbacks compared to conventional images. 
\end{itemize}

\section{Related work}
\label{sec:related}

\subsection{Event Camera Applications}

As previously mentioned, event cameras provide valuable advantages over conventional cameras in many situations. We find recent works which have proved these advantages in several tasks typically solved with conventional vision sensors. Most of these works focus their efforts on 3D reconstruction \cite{rebecq2017emvsevent, kim2016realevent, zhu2018realtimeevent, zhou2018semievent} and 6-DOF camera tracking \cite{rebecq2017evoevent, gallego2018event}. Although 3D reconstruction and localization solutions are very mature on RGB images, existing algorithms cannot be applied exactly the same way on event cameras. The aforementioned works propose different approaches for adapting them.

We find recent approaches that explore the use of these cameras for other tasks,
such as optical flow estimation \cite{gallego2018unifying, liu2018adaptive, zhu2018evflownet} or, closer to our target tasks, object detection and recognition \cite{orchard2015hfirst, chen2018pseudo, lagorce2017hots, sironi2018hats}. Regarding the data used in these recognition works,  Orchard et al. \cite{orchard2015hfirst} and Lagorce et al. \cite{lagorce2017hots} performed the recognition task on small datasets, detecting mainly characters and numbers. %In contrast, 
The most recent works, start to use more challenging (but scarce) recordings in real scenarios, such as N-CARS dataset, used in Sironi et al. \cite{sironi2018hats}, or DDD17 dataset \cite{binas2017ddd17}, which we use in this work because of the real world urban scenarios it contains.

Most of these approaches have a common first step: encode the event information into an image-like representation, in order to facilitate its processing. 
We discuss in detail different previous work event representations (encoding spatial and sometimes temporal information) as well as our proposed representation (with a different way of encoding the temporal information) in Sec. \ref{sec:events-to-segmentation}.

\subsection{Semantic Segmentation}
Semantic segmentation is a visual recognition problem which consists of assigning a semantic label to each pixel in the image.
State-of-the-art on this problem is currently achieved by deep learning based solutions, most of them proposing different variations of encoder-decoder CNN architectures \cite{deeplabv3plus2018, chen2017rethinking, jegou2017one, he2016deep}.

Some of the existing solutions for semantic segmentation target an instance-level semantic segmentation, e.g., Mask-RCNN~\cite{he2017mask}, that includes three main steps: region proposal, binary segmentation and classification. Other solutions, such as  DeepLabv3+~\cite{deeplabv3plus2018}, target class-level semantic segmentation. Deeplabv3+ is a fully convolutional extension of Xception~\cite{Chollet2017XceptionDL}, which is also a state-of-the-art architecture for image classification and the base architecture of our work. 
A survey on image segmentation by Zhu et al. \cite{zhu2016beyond} provides a detailed compilation of more conventional solutions for semantic segmentation, while Garcia et al. \cite{garcia2017review} present a discussion of more recent deep learning based approaches for semantic segmentation, covering from new architectures to common datasets. 

The works discussed so far show the effectiveness of CNNs for semantic segmentation using \textit{RGB} images. 
Closer to our work, we find additional works which prove great performance in semantic segmentation tasks using additional input data modalities to the standard RGB image. 
For example, a common additional input data for semantic segmentation is depth information. Cao et al. \cite{cao2017exploiting} and Gupta et al. \cite{gupta2014learning} are two good examples of how to combine \textit{RGB} images  with  depth information using convolutional neural networks. Similarly, a very common sensor in the robotic field, the Lidar sensor, has also been shown to provide useful additional information when performing semantic segmentation \cite{sun2018developing, dechesne2017semantic}. 
Other works show how to combine less frequent modalities such as fluorescence information  \cite{alonso2017coral} or how to perform semantic segmentation on multi-spectral images \cite{dechesne2017semantic}. 
Semantic segmentation tasks for medical image analysis \cite{litjens2017survey} also typically apply or adapt CNN based approaches designed for RGB images to different medical imaging sensors, such as MRI images \cite{kayalibay2017cnn, milletari2016v} and CT data \cite{christ2016automatic}. 

Our work is focused on a different modality, event camera data, not explored in prior work for semantic segmentation. 
Following one of the top performing models on semantic segmentation for RGB images \cite{deeplabv3plus2018},
we base our network on the Xception design \cite{Chollet2017XceptionDL} to build an encoder-decoder  architecture for semantic segmentation on event images.
Our experiments show good semantic segmentation results using only event data from a public benchmark~\cite{binas2017ddd17}, close to what is achieved on standard imagery from the same scenarios. We also demonstrate the complementary benefits that this modality can bring when combined with standard cameras to solve this problem more accurately.

\section{From Events to Semantic Segmentation}
\label{sec:events-to-segmentation}

\subsection{Event Representation}
\label{sec:event-representation}

\begin{figure*}[!tb]
\centering
    \begin{tabular}{c@{\hspace{1mm}}c@{\hspace{1mm}}c@{\hspace{1mm}}c@{\hspace{1mm}}c@{\hspace{1mm}}}
    Grayscale  & $Hist(x, y, -1)$  &$S(x, y, -1)$ &$Recent(x,y,-1)$ &  $M(x, y, -1)$ \\
    \includegraphics[width=.18\linewidth]{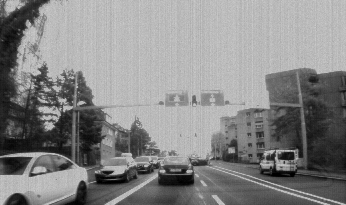} &
        \includegraphics[width=.18\linewidth]{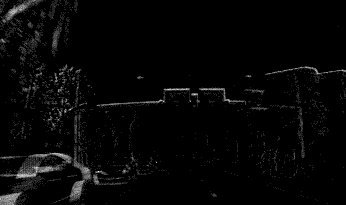}&
        \includegraphics[width=.18\linewidth]{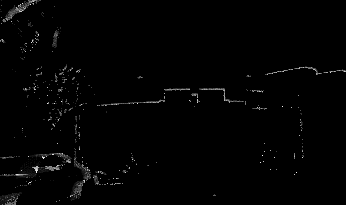}&
        
         \includegraphics[width=.18\linewidth]{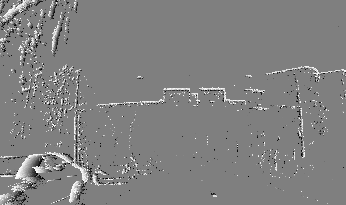} &
         \includegraphics[width=.18\linewidth]{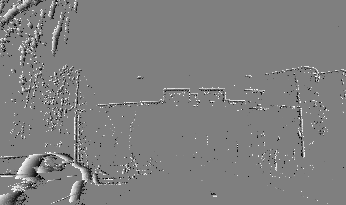} 
    \end{tabular}
  \caption{Visualization (between 0 and 255 gray values) of different 1-channel encodings of data from events with negative polarity ($p = -1)$. 
  In these examples the event information has been integrated for a time interval of 50ms ($T=50ms$). Grayscale is shown as reference.}
\label{fig:representations}
\end{figure*}

Event cameras are very different from conventional \textit{RGB} cameras. Instead of encoding the appearance of the scene within three color channels, they only capture the changes in intensities for each pixel. 
The output of an event camera is not a 3-dimensional image (height, width and channels) but a stream of events. 
An event represents the positive or negative change in the log of the intensity signal (over an established threshold $\sigma$):
\begin{equation}
\log(I_{t+1}) - \log(I_{t}) \geq \sigma,
\label{eq:event_description}
\end{equation}
being $I_{t+1}$ and $I_{t}$ the intensity  captured at two consecutive timestamps. 

Each event ($e_i$) is then defined by four different components: two coordinates ($x_i, y_i$) of the pixel where the change has been measured, a polarity ($p_i$) that can be positive or negative, and a timestamp ($t_i$):
\begin{equation}
e_i = \big\{ x_i, y_i, p_i, t_i \big\}.
\label{eq:event_description2}
\end{equation}
Note there is no representation of the absolute value of the intensity change, only its location and direction (positive polarity, $p=1$, and negative polarity, $p=-1$).

Events are asynchronous and have the described specific encoding that, by construction, does not provide a good input for broadly used techniques in visual recognition tasks nowadays, such as CNNs.
Perhaps the most straightforward representation would be a $nx4$ matrix, with $n$ the number of events. But obviously this representation does not encode the spatial  relationship between events. 
Several strategies have been proposed to encode this information into a dense representation successfully applied in different applications.

\paragraph{Basic dense encoding of event location.}
The most successfully applied event data representation creates a image with several channels encoding the following information. It stores at each location $(x_i, y_i)$ information from the events that happened there at any time $t_i$ within an established integration interval of size $T$. Variations of this representation have been used by many previous works, showing great performance in very different applications: optical flow estimation~\cite{zhu2018evflownet}, object detection \cite{chen2018pseudo}, classification \cite{lagorce2017hots, park2016performance, sironi2018hats} and regression tasks \cite{maqueda2018event}, respectively. 

Earlier works used only one channel to encode event occurrences. Nguyen et al. \cite{nguyen2017real} stores the information of the last event that has occurred in each pixel, i.e., the corresponding value chosen to represent a positive event, negative event or absence of events. One important drawback is that only the last event information remains.
 
In a more complete representation, a recent work for steering wheel angle estimation, from Maqueda et al. \cite{maqueda2018event}, stores the positive and negative event occurrences into two different channels. In other words, this representation ($Hist$) encodes the 2D histogram of positive and negative events %$e_i$ 
that occurred at each pixel $(x_i, y_i)$, as follows: 
\begin{equation}
Hist(x, y, p)=  \sum_{i=1, t_i \epsilon W }^N  \delta(x_i, x)\delta(y_i, y)\delta(p_i, p),
\label{eq:histogram}
\end{equation}% 
where $\delta$ is the Kronecker delta function (the function is 1 if the variables are equal, and 0 otherwise), $W$ %T 
is the time window, or interval, considered to aggregate the event information, and $N$ is the number of events occurred within interval $W$. 
Therefore, the multiplication $\delta(x_i, x)\delta(y_i, y)\delta(p_i, p)$ denotes whether an event $e_i$ matches its coordinates $x_i, y_i$ with $x, y$  values and its polarity $p_i$ with $p$. 
This representation  has two channels, one per polarity $p$ (positive and negative events). 

Note that all the representations discussed so far only use the temporal information (timestamps  $t_i$) to see the time interval where each event belongs to.

\paragraph{Dense encodings including temporal information.}
However, temporal information, i.e., the timestamp of each event $t_i$, contains useful information for recognition tasks, and it has been shown that including this non-spatial information of each event into the image-like encodings is useful. Lagorce et al. \cite{lagorce2017hots} propose a 2-channel image, one channel per polarity, called \textit{time surfaces}. They store, for each pixel, information relative only to the most recent event timestamp during the integration interval $W$. Later, Sironi et al. \cite{sironi2018hats} enhance this previous representation by changing the definition of the \textit{time surfaces}. They now compute the value for each pixel combining information from all the timestamps of events that occurred within $W$. 

Another recently proposed approach by Zhu et al. \cite{zhu2018evflownet} introduces a more complete representation that includes both channels of event occurrence histograms from Maqueda et al. \cite{maqueda2018event}, and two more channels containing temporal information. These two channels ($Recent$) store, at each pixel $(x_i, y_i)$, the normalized timestamp of the most recent positive or negative event, respectively, that occurred in that location during the integration interval: 
\begin{equation}
Recent(x, y, p)= \max\limits_{t_i \epsilon W } t_i \delta(x_i, x)\delta(y_i, y)\delta(p_i, p).
\label{eq:recent}
\end{equation}

All these recent representations  normalize the event timestamps and histograms to be relative values within the interval $W$.\\
Inspired by all this prior work, we propose an alternative representation that combines the best ideas demonstrated so far: the 2-channels of event histograms to encode the spatial distribution of events, together with information regarding all timestamps occurred during integration interval.

Our event representation is a 6-channel image. The first two channels are the histogram of positive and negative events (eq. \ref{eq:histogram}). 
The remaining four channels are a simple but effective way to store information relative to all event timestamps happening during interval $W$. We could see it as a way to store how they are distributed along $T$ rather than selecting just one of the timestamps. 
We propose to store the mean ($M$) and standard deviation ($S$) of the normalized timestamps of events happening at each pixel $(x_i, y_i)$, computed separately for the positive and negative events, as follows: 
\begin{equation}
\small{
M(x, y, p)=  \frac{1}{Hist(x, y, p)} \sum_{i=1, t_i \epsilon W }^N  t_i \delta(x_i, x)\delta(y_i, y)\delta(p_i, p)},
\label{eq:mean}
\end{equation}

\begin{equation}
\scriptsize{
S(x, y, p)=  \sqrt {\ddfrac {\sum_{i=1, t_i \epsilon W }^{N}(t_i \delta(x_i, x)\delta(y_i, y)\delta(p_i, p) - {Mean(x, y, p)})^{2}}{Hist(x, y, p) - 1}}.}
\label{eq:std}
\end{equation}

Then, our representation consists of these six 2D-channels : $Hist(x, y, -1)$, $Hist(x, y, +1)$, $M(x, y, -1)$, $M(x, y, +1)$, $S(x, y, -1)$, $S(x, y, +1)$. 
Figure \ref{fig:representations} shows a visualization of some of these channels. In the event representation images, the brighter the pixels, the higher the value encoded, e.g., white means the highest number of negative events in the $Hist(x, y, -1)$.

\subsection{Semantic Segmentation from Event Data}
\label{sec:approach}

\begin{figure}[!tb]
\centering
\includegraphics[width=1\linewidth]{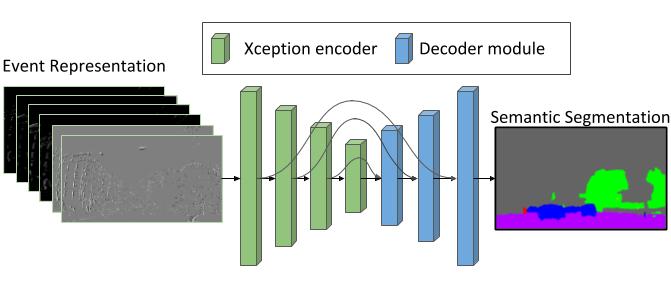}
\caption{Semantic segmentation from event based cameras. We process the different 2D event-data encodings with our encoder-decoder architecture based on Xception \cite{Chollet2017XceptionDL} (Sec. \ref{sec:approach} for more details). Best viewed in color.}%
\label{fig:architecture}
\end{figure}

CNNs have already been shown to work well on dense event-data representations, detailed in previous section \cite{maqueda2018event, zhu2018evflownet}, therefore we explore a CNN based architecture to learn a different visual task, semantic segmentation.
Semantic segmentation is often modelled as a per-pixel classification, and therefore the output of semantic segmentation models has the same resolution that the input. As previously mentioned, there are plenty of recent successful CNN-based approaches to solve this problem both using RGB data and additional modalities. 
We have built an architecture inspired on current state-of-the-art semantic segmentation CNNs, slightly adapted to use the event data encodings.  
Related works commonly follow an encoder-decoder architecture, as we do. As the encoder, we use the well-known Xception model \cite{Chollet2017XceptionDL}, which has been shown to outperform other encoders, both in classification \cite{Chollet2017XceptionDL} and semantic segmentation tasks~\cite{deeplabv3plus2018}. As the decoder, also following state-of-the-art works \cite{chen2017rethinking, deeplabv3plus2018}, we build a \textit{light} decoder, concentrating the heavy computation on the encoder. 
Our architecture also includes features from the most successful recent models for semantic segmentation, including: the use of skip connections to help the optimization of deep architectures \cite{he2016deep, jegou2017one} to avoid the vanishing gradient problem and the use of an auxiliary loss \cite{zhao2017pyramid} which also improves the convergence of the learning process.
Fig. \ref{fig:architecture} shows a diagram of the architecture built in this work, with the multi-channel event representation as network input. 

\begin{table*}[!tb]
    \centering
    \footnotesize
    \begin{tabular}{cccc}
    \hline
      \textbf{Dataset Classes:}   &  \multicolumn{2}{l}{\textit{flat} (road and pavement), \textit{background} (construction and sky), \textit{object}, \textit{vegetation}, \textit{human}, \textit{vehicle}} \\
      \hline
      \textbf{Train Sequences}  & Selected suitable sequence intervals  & Num. Frames   \\
       1487339175 & [0, 4150), [5200, 6600) & 5550 \\
       1487842276 & [1310, 1400), [1900, 2000), [2600, 3550) & 1140  \\
       1487593224 & [870, 2190) & 995  \\
       1487846842 & [380, 500), [1800, 2150), [2575, 2730), [3530, 3900) & 1320  \\

        1487779465 & [1800, 3400), [4000, 4700), [8400, 8630), [8800, 9160), [9920, 10175), [18500, 22300)  & 6945  \\

         &             \multicolumn{1}{r}{\textbf{TOTAL:}}             & \textbf{15950} \\
         \hline
       \textbf{Test Sequences}  & Selected suitable sequence intervals  & Num. Frames   \\
      1487417411 &  [100, 1500), [2150, 3100), [3200, 4430), [4840, 5150) & \textbf{3890}   \\ 
       \hline
    \end{tabular}
    \caption{Summary of Ev-Seg Data which consists of several intervals of some sequences of the DDD17 dataset.}
    \label{tab:my_label}
\end{table*}

As similar architectures, we perform the training optimization via back-propagation of the loss,  calculated as the summation of all per-pixel losses, through the parameter gradients. 
We use the common soft-max cross entropy loss function ($\mathcal{L}$) described in eq. (\ref{eq:lossf}):
\begin{equation}
\mathcal{L}= -\frac{1}{N}\sum_{j=1}^N\sum_{c=1}^My_{c,j}\ln(\hat{y}_{c,j}),
\label{eq:lossf}
\end{equation}%
\noindent where $N$ is the number of labeled pixels and \textit{M} is the number of classes. ${Y}_{c,j}$ is a binary indicator of pixel $j$  belonging to class \textit{c} (ground truth). $\hat{Y}_{c,j}$ is the CNN predicted probability of pixel $j$ belonging to class \textit{c}.

\section{Ev-Seg: Event-Segmentation Data}
\label{sec:data}

The Ev-Seg data is an extension for semantic segmentation of the DDD17 dataset \cite{binas2017ddd17} (which does not provide semantic segmentation labels). Our extension includes generated (automatically generated, non-manual annotations) semantic segmentation labels to be used as ground truth for a large subset of that dataset.
Besides the labels, to facilitate replication and further experimentation, together with the labeling, we also publish the selected subset of  grayscale images and corresponding event data encoded with three different representations (Maqueda et al. \cite{maqueda2018event}, Zhu et al.  \cite{zhu2018evflownet} and the new one proposed in this work).

\paragraph{Generating the labels.} 
 Besides the obvious burden of manually labeling a semantic segmentation per-pixel ground truth, if we think of performing this task directly on the event-based data it turns out even more challenging. We only need to look at any of the event representations available (see Fig. \ref{fig:introduction-results}), to realize that for the human eye is hard to distinguish many of the classes there if the grayscale image is not side-by-side. 
Other works have shown how CNNs are robust to training with noise~\cite{sun2017revisiting} or approximated labels\cite{alonso2017coral}, including the work of Chen et al. \cite{chen2018pseudo} that also succesfully uses generated labels from grayscale for object detection in event-based data. 
We then propose to use the corresponding grayscale images to generate an approximated set of labels for training, which we demonstrate is enough to train models to segment directly on event based data.

To generate these approximated semantic labels, we performed the following three steps.

First, we have trained a CNN for semantic segmentation on the well known urban environment dataset \textit{Cityscapes}~\cite{cordts2016cityscapes}, but using grayscale version of all its images. The architecture used for this step is the same architecture described in subsection \ref{sec:approach}, which follows state-of-the-art components for semantic segmentation. This grayscale segmentation model was trained for 70 epochs with a learning rate of 1e-4. The final model obtains 83\% categories MIoU on the \textit{Cityscapes} validation data. This is still a bit far from the top results obtained on that dataset with RGB images (92\% MIoU), but enough quality for our the process.

Secondly, with this grayscale model, we obtained the semantic segmentation on all grayscale images of the selected sequences (we detail next which sequences were used and why). These segmentations are what we will consider the labels to train our event-based segmentation model. 

Lastly, as a final post-processing step on the ground truth labels, we cropped the bottom part of all the images, i.e., 60 bottom rows of the image it always contains the car dashboard and it only introduces noise into the generated labels.

\paragraph{Subset of DDD17 sequences selection.} 
\label{sec:data-generation}
As previously mentioned, we have not generated the labels for all the DDD17 data. We next discuss the reasons and selection criteria that we followed.

The DDD17 dataset consists of 40 sequences of different driving set-ups. These sequences were recorded on different scenarios (e.g., motorways and urban scenarios), with very different illumination conditions: some of them have been recorded during day-time (where everything is clear and visible), but others have overexposure or have been recorded at night, making some of the grayscale images almost useless for standard visual recognition approaches.

As the data domain available to train the base grayscale semantic segmentation model was \textit{Cityscapes} data, which is a urban domain, we selected only the sequences from urban scenarios. Besides, only images with enough contrast (not too bright, not too dark) are likely to provide a good generated ground truth. Therefore, we only selected sequences which were recorded during day-time, with no extreme overexposure. Given these restrictions, only six sequences approximately matched them. Therefore,  we performed a manual more detailed annotation of the intervals in each of these sequences where the restrictions applied (details on Table \ref{tab:my_label}).

\begin{figure}[!t]
\centering
    \begin{tabular}{c@{\hspace{1mm}}c@{\hspace{1mm}}}

    Grayscale  &   Label  \\
    \includegraphics[width=.41\linewidth]{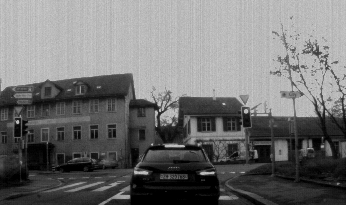} &
    \includegraphics[width=.41\linewidth]{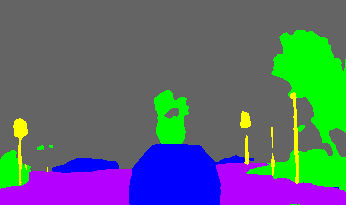} \\
        \includegraphics[width=.41\linewidth]{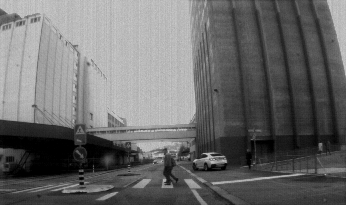}&
    \includegraphics[width=.41\linewidth]{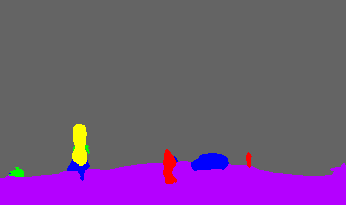} \\
        \includegraphics[width=.41\linewidth]{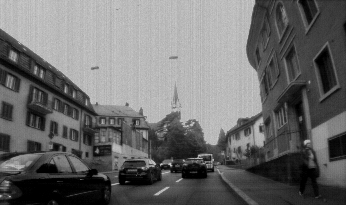} &
    \includegraphics[width=.41\linewidth]{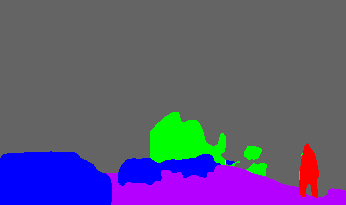} \\
    \end{tabular}
  \caption{Three examples of the Ev-Seg data generated for the test sequence. Semantic label images (right) have been generated from the grayscale images (left) through a CNN trained on a grayscale version of Cityscapes. Best viewed in color.}
\label{fig:examples_dataset}
\end{figure}

\paragraph{Data summary.} 
Table  \ref{tab:my_label} shows a summary of the contents of the Ev-Seg data.   
From the six sequences selected as detailed previously, five sequences were used as training data and one sequence was used for testing.  
We chose for testing  the sequence with more homogeneous class distribution, i.e., that contained  more amount of labels of categories which appears less such as the human/pedestrian label.

The labels have the same categories than the well-known \textit{Cityscapes} dataset \cite{cordts2016cityscapes} (see Table \ref{tab:my_label}), with the exception of \textit{sky} and \textit{construction} categories. Although these two categories  were properly learned in the Cityscapes dataset, when performing inferences on the DDD17 dataset with grayscale images, these categories were not correctly generated due to the domain-shift. Therefore in our experiments, those two categories are learned together, as if they were the same thing.
This domain shift between the Cityscapes and DDD17 datasets was also the cause of generating the Cityscapes categories in stead of its classes. 

 Figure \ref{fig:examples_dataset} shows three  examples of grayscale images and corresponding generated segmentation that belong to our extension of the \textit{DDD17 dataset.}
 We can see that although the labels are not as perfect as if manually annotated (and as previously mentioned, classes such as \textit{building} and \textit{sky} were not properly learned using only grayscale), they are pretty accurate and well defined. 
 
 \begin{table*}[!tb]
\begin{center}
\begin{tabular}{|l||c|c||c|c||c|c|}
\hline
 & Accuracy  & MIoU   & Accuracy  & MIoU  & Accuracy   & MIoU  \\
Input representation &  50ms &  50ms  &  10ms &  10ms &  250ms  &  250ms \\
\hline\hline
Maqueda et al. \cite{maqueda2018event} & 88.85  & 53.07  & 85.06& 42.93  & 87.09& 45.66 \\
Zhu et al. \cite{zhu2018evflownet} & 88.99 & 52.32 & 86.35& 43.65  & 85.89 & 45.12 \\
Ours & \textbf{89.76} & \textbf{54.81}  & \textbf{ 86.46} & \textbf{45.85 }& \textbf{87.72}& \textbf{47.56} \\
\hline
Grayscale  & 94.67 & 64.98  &  94.67 & 64.98 &  94.67 & 64.98  \\
Grayscale \& Ours  &\textbf{95.22}  & \textbf{68.36} &  \textbf{95.18} &  \textbf{67.95} & \textbf{95.29}  & \textbf{68.26} \\
\hline
\end{tabular}
\end{center}
\caption{Semantic segmentation performance of different input representations on the test Ev-Seg data.  Models trained using time intervals ($T$) of 50ms but tested with different $T$ values: 50ms, 10ms and 250ms. }
\label{tab:results}
\end{table*}

\section{Experimental Validation}
\label{sec:experiments}

\begin{figure*}[!tb]
\centering
    \small
    \begin{tabular}{c@{\hspace{2mm}}c@{\hspace{1mm}}c@{\hspace{1mm}}c@{\hspace{2mm}}c@{\hspace{1mm}}c@{\hspace{2mm}}c@{\hspace{1mm}}}

    Grayscale img & \footnotesize{Maqueda et al. \cite{maqueda2018event}  } &  Zhu et al. \cite{zhu2018evflownet}  & \textbf{Ours} &  Grayscale   &   \footnotesize{\textbf{Grayscale \& Ours}} & GT Labels  \\

    \includegraphics[width=.13\linewidth]{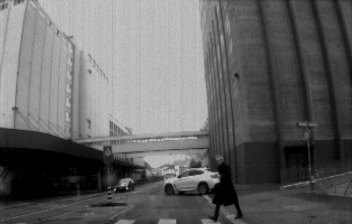} 
    &\includegraphics[width=.13\linewidth]{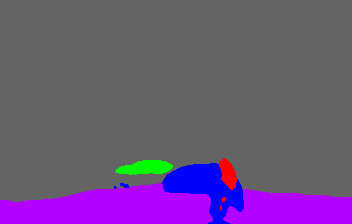} 
    &\includegraphics[width=.13\linewidth]{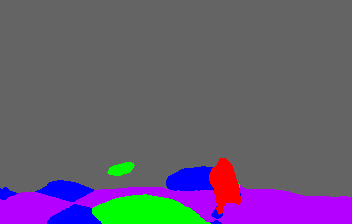} 
    &\includegraphics[width=.13\linewidth]{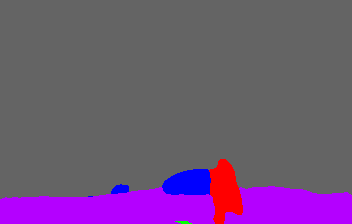} 
    &\includegraphics[width=.13\linewidth]{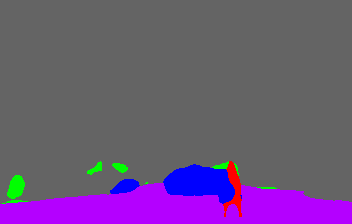} 
    &\includegraphics[width=.13\linewidth]{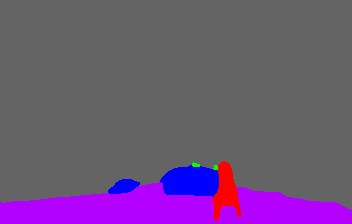} 
    &\includegraphics[width=.13\linewidth]{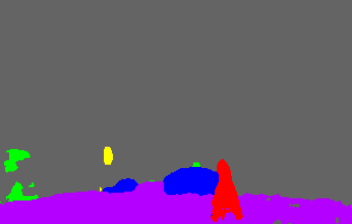} 
    \\

    \includegraphics[width=.13\linewidth]{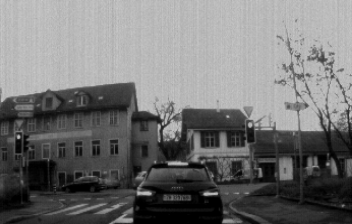} 
    &\includegraphics[width=.13\linewidth]{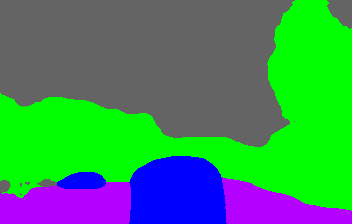} 
    &\includegraphics[width=.13\linewidth]{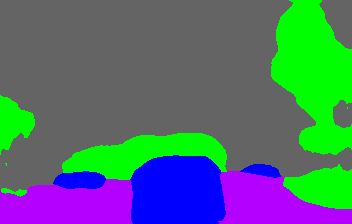} 
    &\includegraphics[width=.13\linewidth]{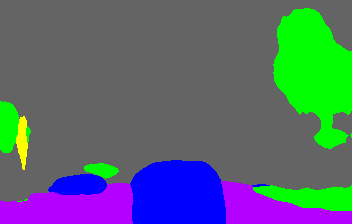} 
    &\includegraphics[width=.13\linewidth]{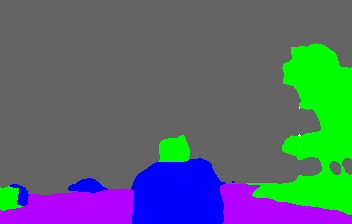} 
    &\includegraphics[width=.13\linewidth]{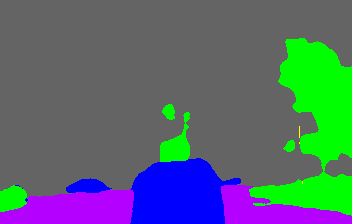} 
    &\includegraphics[width=.13\linewidth]{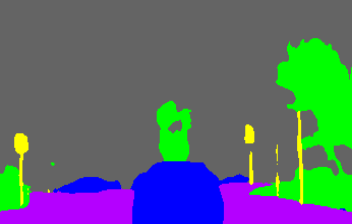} 
    \\

    \includegraphics[width=.13\linewidth]{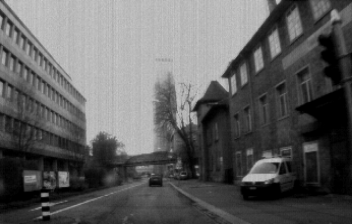} 
    &\includegraphics[width=.13\linewidth]{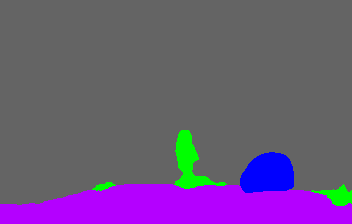} 
    &\includegraphics[width=.13\linewidth]{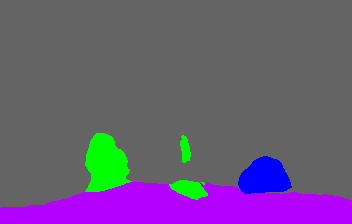} 
    &\includegraphics[width=.13\linewidth]{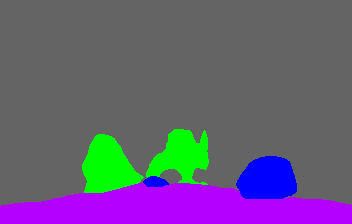} 
    &\includegraphics[width=.13\linewidth]{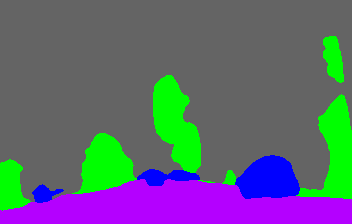} 
    &\includegraphics[width=.13\linewidth]{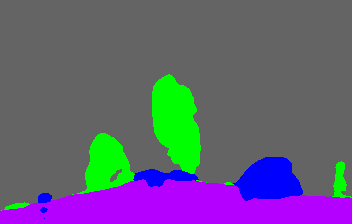} 
    &\includegraphics[width=.13\linewidth]{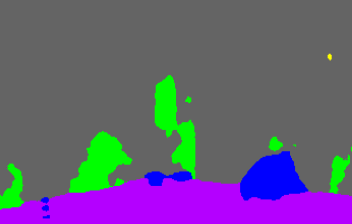} 
    \\
   \includegraphics[width=.13\linewidth]{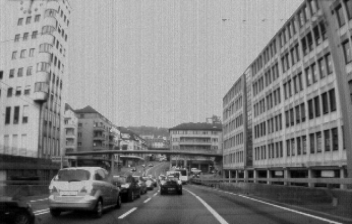} &
   \includegraphics[width=.13\linewidth]{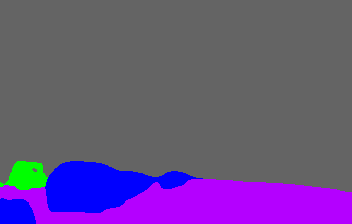} &
   \includegraphics[width=.13\linewidth]{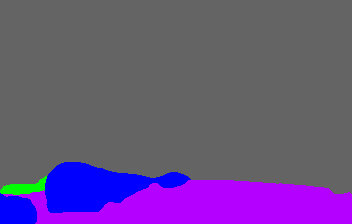} &
   \includegraphics[width=.13\linewidth]{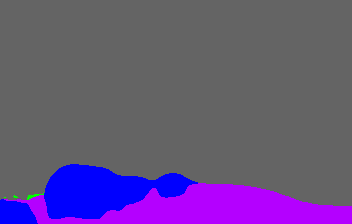} &
   \includegraphics[width=.13\linewidth]{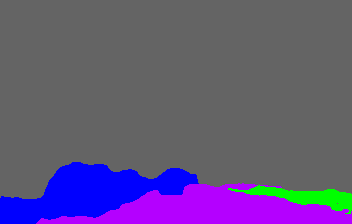} &
   \includegraphics[width=.13\linewidth]{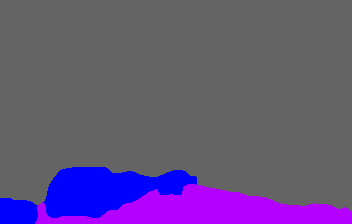} &
   \includegraphics[width=.13\linewidth]{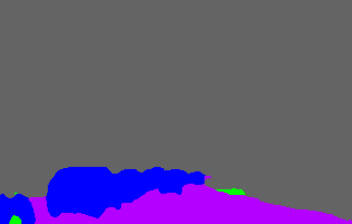}
   \\
    \includegraphics[width=.13\linewidth]{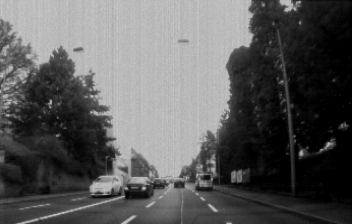} 
    &\includegraphics[width=.13\linewidth]{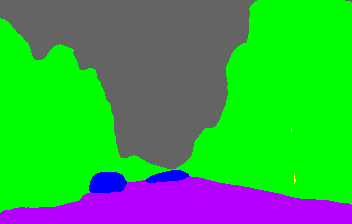} 
    &\includegraphics[width=.13\linewidth]{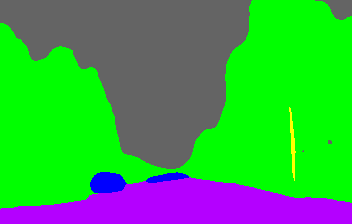} 
    &\includegraphics[width=.13\linewidth]{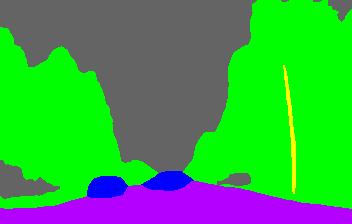} 
    &\includegraphics[width=.13\linewidth]{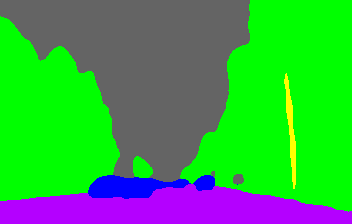} 
    &\includegraphics[width=.13\linewidth]{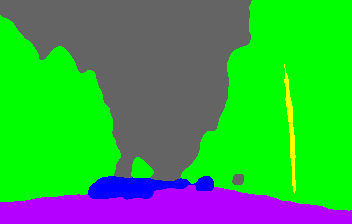} 
    &\includegraphics[width=.13\linewidth]{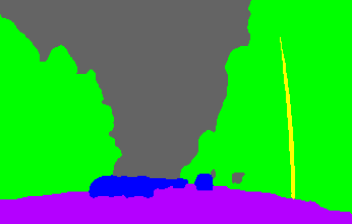} 
    \\
    \includegraphics[width=.13\linewidth]{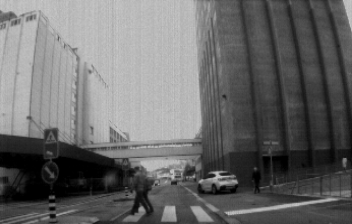} 
    &\includegraphics[width=.13\linewidth]{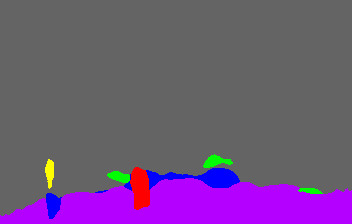} 
    &\includegraphics[width=.13\linewidth]{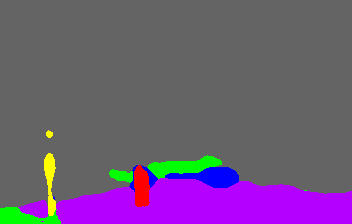} 
    &\includegraphics[width=.13\linewidth]{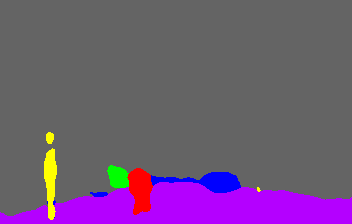} 
    &\includegraphics[width=.13\linewidth]{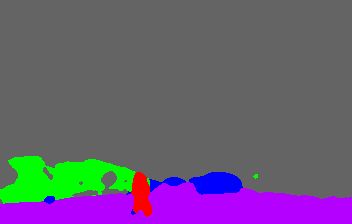} 
    &\includegraphics[width=.13\linewidth]{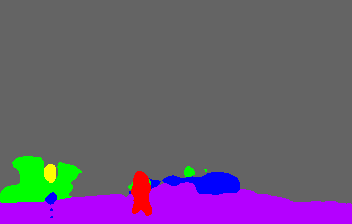} 
    &\includegraphics[width=.13\linewidth]{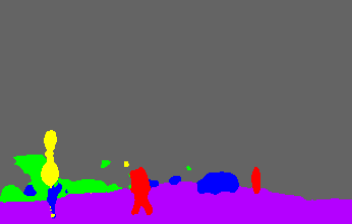} 
    \\
    (a) & (b) & (c) & (d) & (e) & (f) & (g)\\
    \end{tabular}
  \caption{Semantic segmentation on several test images from Ev-Seg data. Results using different input representations of event data only, (b) to (d), or using grayscale data (e) and (f). Grayscale original image (a) and ground truth labels are shown for visualization purposes.  Models trained and tested on time intervals of 50ms. Best viewed in color. 
  }
\label{fig:segmentations}
\end{figure*}

\subsection{Experiment Set-up and Metrics}
\paragraph{Metrics.}
Our work addresses the semantic segmentation problem, i.e., per pixel classification, using event cameras. Thus, we  evaluate our results on the standard metrics for classification and semantic segmentation: \textit{Accuracy} and  \textit{Mean Intersection over Union (MIoU) }.

In semantic segmentation, given a predicted image $\hat{y}$ and a ground-truth image $y$, and being $N$ their number of pixels, which can be classified in $C$ different classes, the accuracy metric, eq. (\ref{eq:accuracy}) is computed as:
\begin{equation}
Accuracy(y,\hat{y})= \frac{1}{N}\sum_{i=1}^N\delta(y_{i}, \hat{y_{i}}),
\label{eq:accuracy}
\end{equation}%
and the MIoU is calculated per class as:
\begin{equation}
\small{
MIoU(y,\hat{y})=\frac{1}{C}\sum_{j=1}^C  \frac{\sum_{i=1}^N  \delta(y_{i, c}, 1)\delta(y_{i, c},  \hat{y}_{i, c})
}{\sum_{i=1}^N max(1, \delta(y_{i, c}, 1)+\delta(\hat{y}_{i, c}, 1))},
\label{eq:miou}
}
\end{equation}
where $\delta$ denotes the Kronecker delta function, $y_{i}$ indicates the class where pixel $i$ belongs to, and $y_{i, c}$ is a boolean that indicates if pixel $i$ belongs to a certain class $c$. %(1 if true, 0 if false).

\paragraph{Set-up.}
We perform the experiments using the CNN explained in Sec. \ref{sec:approach}. and the Ev-Seg data detailed in Sec. \ref{sec:data}. 
We train all model variations from scratch using: the Adam optimizer with an initial learning rate of $1e-4$ and a polynomial learning rate decay schedule. We train for $30K$ iterations using a batch size of 8 and during training we perform several data augmentation steps: crops, rotations (-15$^\circ$, 15$^\circ$), vertical and horizontal shifts (-25\%, 25\%) and horizontal flips.
Regarding the event information encoding, for training we always use an integration time interval $T= 50ms$ which has been shown to perform well on this dataset \cite{maqueda2018event}.

\subsection{Event Semantic Segmentation}

\paragraph{Input representation comparison.}
A good input representation is very important for a CNN to properly learn and exploit the input information.
Table \ref{tab:results} compares several semantic segmentation models trained with different input representations. The top three rows correspond to event-based representations. We compare a basic dense encoding of event locations, a dense encoding which also includes temporal information and our proposed encoding (see Sec.\ref{sec:event-representation}. for details). 
Our event encoding performs slightly but consistently better on the semantic segmentation task on the different metrics and evaluations considered. Fig. \ref{fig:segmentations} shows a few visual examples of these results.

\begin{figure}[!tb]
\centering
    \begin{tabular}{@{}c@{\hspace{1mm}}c@{\hspace{1mm}}c}
    $T=250ms$  & $T=50ms$   & $T=10ms$  \\
    \includegraphics[width=.32\linewidth]{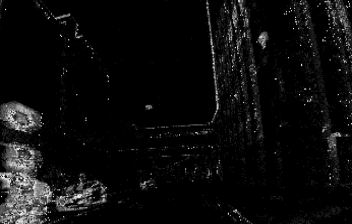} &
    \includegraphics[width=.32\linewidth]{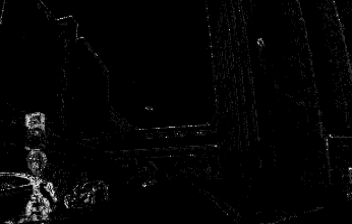} &
    \includegraphics[width=.32\linewidth]{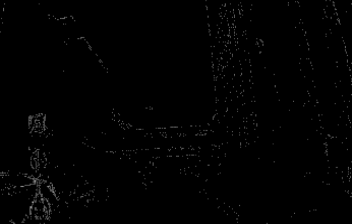}  \\
    \includegraphics[width=.32\linewidth]{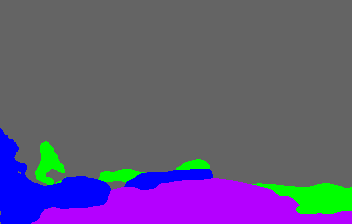}&
        \includegraphics[width=.32\linewidth]{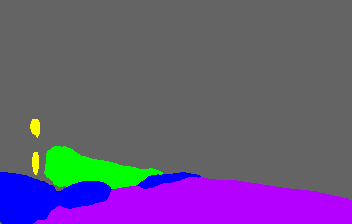} &
    \includegraphics[width=.32\linewidth]{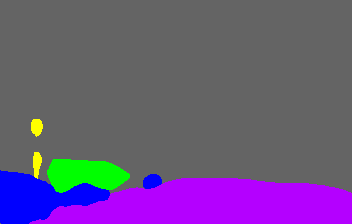} \\
    \end{tabular}
  \caption{Semantic segmentation results (bottom) using different integration interval size ($T)$ for the event data representation (top). Results obtained with a model trained only on 50ms integrated event information encoded with our proposed representation.  Best viewed in color.  }
\label{fig:multiscale-segmentation}
\end{figure}

All models (same architecture, just trained with different inputs) have been trained with data encoded using integration intervals of 50ms, but we also evaluate them using different interval sizes. 
This is an interesting evaluation because by changing the time interval, in which the event information is aggregated, we somehow simulate different camera movement speeds. In other words, intervals of 50ms or 10ms may encode exactly the same movement but at different speeds. This point is pretty important because in real scenarios, models have to perform well at different speeds. 
We can see that all models perform just slightly worse on test data encoded with different intervals sizes (10ms, 250ms) that the integration time used during training (50ms), see Fig. \ref{fig:multiscale-segmentation} examples. There are two main explanations for why the models are performing similar on different integration intervals: 1) the encodings are normalized and 2) the training data contains different camera speeds. Both things help to generalize better on different time intervals or movement speeds. \footnote{The code and data to replicate these experiments will be soon released.}

\paragraph{Event vs conventional cameras.}
Table \ref{tab:results} also includes, in the two bottom rows, results using the corresponding grayscale image for the semantic segmentation task.

Although conventional cameras capture richer pure appearance information than event cameras, event cameras provide motion information, which is also very useful for the semantic segmentation task. 
In examples of results using grayscale data from Fig.~\ref{fig:segmentations}(e), (f), we can see how event information helps for example to better segment moving objects, such as pedestrians (in red in those examples) or to refine object borders.
While conventional cameras suffer detecting small objects and in general, with any recognition on extreme illumination (bright or dark) conditions, event cameras suffer more in recognizing objects with no movement (because they move at the same speed than the camera or because they are too far to appreciate their movement).
See the video supplementary material for the side-by-side segmentation results on complete sequences with the different event-based representations and the conventional camera data.

 Conventional cameras perform better on their own for semantic segmentation than event based cameras on their own. 
 However, our results show that semantic segmenation results are better when combinbing both of them. This suggests they are learning complementary information. Interestingly, we should note that the data available for training and evaluation is precisely data where we could properly segment the grayscale image, therefore slightly more beneficial for grayscale images than event-based data (i.e., there is no night-time image included in the evaluation set because there is no ground truth for those). 
 
\begin{figure}[!tb]
\centering
    \begin{tabular}{c@{\hspace{1mm}}c@{\hspace{1mm}}}

    Grayscale  &   Event representation  \\
    \includegraphics[width=.40\linewidth]{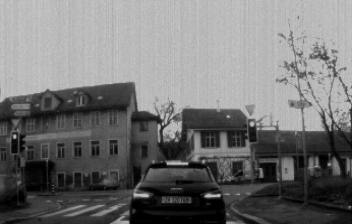} &
    \includegraphics[width=.40\linewidth]{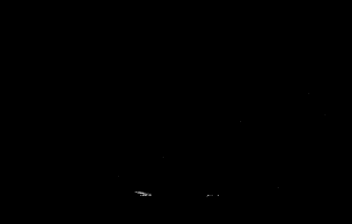}  \\
        \includegraphics[width=.40\linewidth]{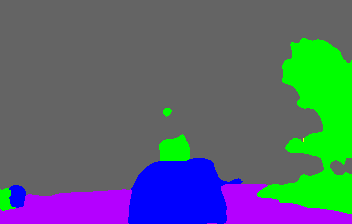} &
    \includegraphics[width=.40\linewidth]{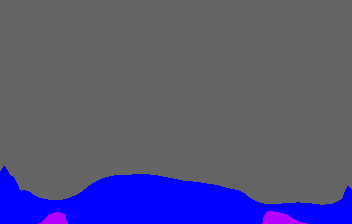} \\
    \end{tabular}
  \caption{Semantic segmentation result (bottom) on a static part of the sequence, i.e, a car waiting at a crossing. This is an obvious adversarial case for event cameras, due to lack of event information. Best viewed in color. }
\label{fig:static}
\end{figure}

\begin{figure}[!tb]
\centering
    \begin{tabular}{c@{\hspace{1mm}}c@{\hspace{1mm}}}

    Grayscale  &   Event representation  \\
    \includegraphics[width=.40\linewidth]{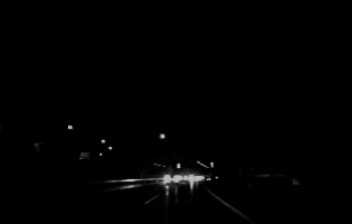} &
    \includegraphics[width=.40\linewidth]{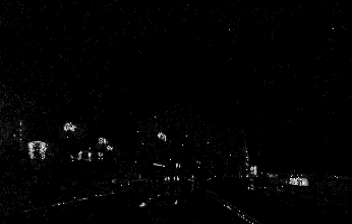}  \\
        \includegraphics[width=.40\linewidth]{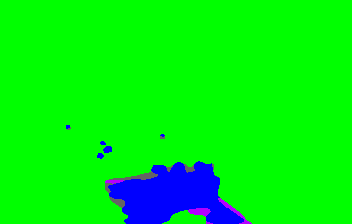} &
    \includegraphics[width=.40\linewidth]{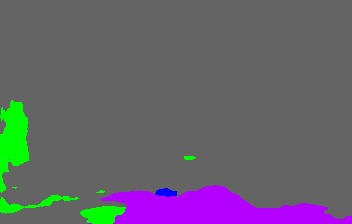} \\
    \end{tabular}
  \caption{Semantic segmentation  (bottom) on extreme lighting conditions (night-time) 
  with different input representations (top): grayscale image and our event data representation. Corresponding models trained only on good illuminated daytime samples. This is an obvious adversarial case for conventional cameras, due to lack of information in the grayscale capture. Best viewed in color.}
\label{fig:night-segmentation}
\end{figure}

Two clear complementary situations from our experiments:
1) On one hand, it is already known that one the major drawback of event cameras is that objects that do not move with respect to the camera do not trigger events, i.e., are invisible. Fig. \ref{fig:static} shows an example of a car waiting at a pedestrian crossing, where we see that while conventional cameras can perfectly see the whole scene, event cameras barely capture any information; 2) On the other hand, event cameras are able to capture meaningful information on situations where scene objects are not visible at all for conventional vision sensors, e.g., difficult lighting environments. This is due to their high dynamic range, Fig. \ref{fig:night-segmentation} illustrates an example of a situation where neither of the grayscale nor the event-based models have been trained for. The event-based model performs much better due to the minor domain-shift on the input representation. 

\section{Conclusions and Future Work}
\label{sec:conclusion}
This work includes the first results on semantic segmentation using event camera information. We build an Xception-based encoder-decoder architecture  which is able to learn semantic segmentation only from event camera data.
Since there is no benchmark available for this problem, we propose how to generate automatic but approximate semantic segmentation labels for some sequences of the DDD17 event-based dataset. 
Our evaluation shows how this approach allows the effective learning of semantic segmentation models from event data. Both models and generated labeled data are being released.

In order to feed the model, we also propose a novel event camera data representation, which encodes both the event histogram and their temporal distribution. 
Our semantic segmentation experiments,  show that our approach outperforms other previously used event representations, even when evaluating in different time intervals. 
We also compare the segmentation achieved only from event data to the segmentation from conventional images, showing their benefits, their drawbacks and the benefits of combining both sensors for this task.

For future work, one of the main challenges is still obtaining and generating more and better semantic segmentation labels, through alternative domain adaptation approaches and/or event camera simulators (they currently do not provide this kind of labels). Besides, it would be also interesting, and not only for the currently explored recognition problem of segmentation, to develop alternative architectures and data augmentation methods more specific for event based cameras.

\section*{Acknowledgements}
The authors would like to thank NVIDIA Corporation for the donation of a Titan Xp GPU used in this work.

{\small
\bibliographystyle{ieee}
\bibliography{main}

\begin{thebibliography}{10}\itemsep=-1pt

\bibitem{alonso2017coral}
I.~Alonso, A.~Cambra, A.~Munoz, T.~Treibitz, and A.~C. Murillo.
\newblock Coral-segmentation: Training dense labeling models with sparse ground
  truth.
\newblock In {\em IEEE Int. Conf. on Computer Vision Workshops}, pages
  2874--2882, 2017.

\bibitem{binas2017ddd17}
J.~Binas, D.~Neil, S.-C. Liu, and T.~Delbruck.
\newblock Ddd17: End-to-end davis driving dataset.
\newblock {\em ICML Workshop on Machine Learning for Autonomous Vehicles},
  2017.

\bibitem{cao2017exploiting}
Y.~Cao, C.~Shen, and H.~T. Shen.
\newblock Exploiting depth from single monocular images for object detection
  and semantic segmentation.
\newblock {\em IEEE Transactions on Image Processing}, 26(2):836--846, 2017.

\bibitem{chen2017rethinking}
L.-C. Chen, G.~Papandreou, F.~Schroff, and H.~Adam.
\newblock Rethinking atrous convolution for semantic image segmentation.
\newblock {\em arXiv preprint arXiv:1706.05587}, 2017.

\bibitem{deeplabv3plus2018}
L.-C. Chen, Y.~Zhu, G.~Papandreou, F.~Schroff, and H.~Adam.
\newblock Encoder-decoder with atrous separable convolution for semantic image
  segmentation.
\newblock In {\em ECCV}, 2018.

\bibitem{chen2018pseudo}
N.~F. Chen.
\newblock Pseudo-labels for supervised learning on dynamic vision sensor data,
  applied to object detection under ego-motion.
\newblock In {\em Proceedings of the IEEE Conference on Computer Vision and
  Pattern Recognition Workshops}, pages 644--653, 2018.

\bibitem{Chollet2017XceptionDL}
F.~Chollet.
\newblock Xception: Deep learning with depthwise separable convolutions.
\newblock {\em 2017 IEEE Conference on Computer Vision and Pattern Recognition
  (CVPR)}, pages 1800--1807, 2017.

\bibitem{christ2016automatic}
P.~F. Christ, M.~E.~A. Elshaer, F.~Ettlinger, S.~Tatavarty, M.~Bickel,
  P.~Bilic, M.~Rempfler, M.~Armbruster, F.~Hofmann, M.~D’Anastasi, et~al.
\newblock Automatic liver and lesion segmentation in ct using cascaded fully
  convolutional neural networks and 3d conditional random fields.
\newblock In {\em International Conference on Medical Image Computing and
  Computer-Assisted Intervention}, pages 415--423. Springer, 2016.

\bibitem{cordts2016cityscapes}
M.~Cordts, M.~Omran, S.~Ramos, T.~Rehfeld, M.~Enzweiler, R.~Benenson,
  U.~Franke, S.~Roth, and B.~Schiele.
\newblock The cityscapes dataset for semantic urban scene understanding.
\newblock In {\em Proc. of IEEE conf. on CVPR}, pages 3213--3223, 2016.

\bibitem{dechesne2017semantic}
C.~Dechesne, C.~Mallet, A.~Le~Bris, and V.~Gouet-Brunet.
\newblock Semantic segmentation of forest stands of pure species combining
  airborne lidar data and very high resolution multispectral imagery.
\newblock {\em ISPRS Journal of Photogrammetry and Remote Sensing},
  126:129--145, 2017.

\bibitem{gallego2018event}
G.~Gallego, J.~E. Lund, E.~Mueggler, H.~Rebecq, T.~Delbruck, and D.~Scaramuzza.
\newblock Event-based, 6-dof camera tracking from photometric depth maps.
\newblock {\em IEEE transactions on pattern analysis and machine intelligence},
  40(10):2402--2412, 2018.

\bibitem{gallego2018unifying}
G.~Gallego, H.~Rebecq, and D.~Scaramuzza.
\newblock A unifying contrast maximization framework for event cameras, with
  applications to motion, depth, and optical flow estimation.
\newblock In {\em IEEE Int. Conf. Comput. Vis. Pattern Recog.(CVPR)}, volume~1,
  2018.

\bibitem{garcia2017review}
A.~Garcia-Garcia, S.~Orts-Escolano, S.~Oprea, V.~Villena-Martinez, and
  J.~Garcia-Rodriguez.
\newblock A review on deep learning techniques applied to semantic
  segmentation.
\newblock {\em arXiv preprint arXiv:1704.06857}, 2017.

\bibitem{gupta2014learning}
S.~Gupta, R.~Girshick, P.~Arbel{\'a}ez, and J.~Malik.
\newblock Learning rich features from rgb-d images for object detection and
  segmentation.
\newblock In {\em European Conference on Computer Vision}, pages 345--360.
  Springer, 2014.

\bibitem{he2017mask}
K.~He, G.~Gkioxari, P.~Doll{\'a}r, and R.~Girshick.
\newblock Mask {R-CNN}.
\newblock In {\em IEEE Int. Conf. on Computer Vision}, pages 2980--2988, 2017.

\bibitem{he2016deep}
K.~He, X.~Zhang, S.~Ren, and J.~Sun.
\newblock Deep residual learning for image recognition.
\newblock In {\em Proc. of IEEE CVPR}, pages 770--778, 2016.

\bibitem{jegou2017one}
S.~J{\'e}gou, M.~Drozdzal, D.~Vazquez, A.~Romero, and Y.~Bengio.
\newblock The one hundred layers tiramisu: Fully convolutional densenets for
  semantic segmentation.
\newblock In {\em CVPRW}, pages 1175--1183. IEEE, 2017.

\bibitem{kayalibay2017cnn}
B.~Kayalibay, G.~Jensen, and P.~van~der Smagt.
\newblock Cnn-based segmentation of medical imaging data.
\newblock {\em arXiv preprint arXiv:1701.03056}, 2017.

\bibitem{kim2016realevent}
H.~Kim, S.~Leutenegger, and A.~J. Davison.
\newblock Real-time 3d reconstruction and 6-dof tracking with an event camera.
\newblock In {\em European Conference on Computer Vision}, pages 349--364.
  Springer, 2016.

\bibitem{lagorce2017hots}
X.~Lagorce, G.~Orchard, F.~Galluppi, B.~E. Shi, and R.~B. Benosman.
\newblock Hots: a hierarchy of event-based time-surfaces for pattern
  recognition.
\newblock {\em IEEE transactions on pattern analysis and machine intelligence},
  39(7):1346--1359, 2017.

\bibitem{lichtsteiner2008128}
P.~Lichtsteiner, C.~Posch, and T.~Delbruck.
\newblock A 128$\times$128 120 db 15$\mu$s latency asynchronous temporal
  contrast vision sensor.
\newblock {\em IEEE journal of solid-state circuits}, 43(2):566--576, 2008.

\bibitem{litjens2017survey}
G.~Litjens, T.~Kooi, B.~E. Bejnordi, A.~A.~A. Setio, F.~Ciompi, M.~Ghafoorian,
  J.~A. Van Der~Laak, B.~Van~Ginneken, and C.~I. S{\'a}nchez.
\newblock A survey on deep learning in medical image analysis.
\newblock {\em Medical image analysis}, 42:60--88, 2017.

\bibitem{liu2018adaptive}
M.~Liu and T.~Delbruck.
\newblock Adaptive time-slice block-matching optical flow algorithm for dynamic
  vision sensors.
\newblock Technical report, 2018.

\bibitem{maqueda2018event}
A.~I. Maqueda, A.~Loquercio, G.~Gallego, N.~Garc{\i}a, and D.~Scaramuzza.
\newblock Event-based vision meets deep learning on steering prediction for
  self-driving cars.
\newblock In {\em Proceedings of the IEEE Conference on Computer Vision and
  Pattern Recognition}, pages 5419--5427, 2018.

\bibitem{milletari2016v}
F.~Milletari, N.~Navab, and S.-A. Ahmadi.
\newblock V-net: Fully convolutional neural networks for volumetric medical
  image segmentation.
\newblock In {\em 3D Vision (3DV), 2016 Fourth International Conference on},
  pages 565--571. IEEE, 2016.

\bibitem{nguyen2017real}
A.~Nguyen, T.-T. Do, D.~G. Caldwell, and N.~G. Tsagarakis.
\newblock Real-time pose estimation for event cameras with stacked spatial lstm
  networks.
\newblock {\em arXiv preprint arXiv:1708.09011}, 2017.

\bibitem{orchard2015hfirst}
G.~Orchard, C.~Meyer, R.~Etienne-Cummings, C.~Posch, N.~Thakor, and
  R.~Benosman.
\newblock Hfirst: a temporal approach to object recognition.
\newblock {\em IEEE transactions on pattern analysis and machine intelligence},
  37(10):2028--2040, 2015.

\bibitem{park2016performance}
P.~K. Park, B.~H. Cho, J.~M. Park, K.~Lee, H.~Y. Kim, H.~A. Kang, H.~G. Lee,
  J.~Woo, Y.~Roh, W.~J. Lee, et~al.
\newblock Performance improvement of deep learning based gesture recognition
  using spatiotemporal demosaicing technique.
\newblock In {\em Image Processing (ICIP), 2016 IEEE International Conference
  on}, pages 1624--1628. IEEE, 2016.

\bibitem{rebecq2017emvsevent}
H.~Rebecq, G.~Gallego, E.~Mueggler, and D.~Scaramuzza.
\newblock Emvs: Event-based multi-view stereo—3d reconstruction with an event
  camera in real-time.
\newblock {\em International Journal of Computer Vision}, pages 1--21, 2017.

\bibitem{rebecq2017evoevent}
H.~Rebecq, T.~Horstschaefer, G.~Gallego, and D.~Scaramuzza.
\newblock Evo: A geometric approach to event-based 6-dof parallel tracking and
  mapping in real time.
\newblock {\em IEEE Robotics and Automation Letters}, 2(2):593--600, 2017.

\bibitem{sironi2018hats}
A.~Sironi, M.~Brambilla, N.~Bourdis, X.~Lagorce, and R.~Benosman.
\newblock Hats: Histograms of averaged time surfaces for robust event-based
  object classification.
\newblock In {\em Proceedings of the IEEE Conference on Computer Vision and
  Pattern Recognition}, pages 1731--1740, 2018.

\bibitem{sun2017revisiting}
C.~Sun, A.~Shrivastava, S.~Singh, and A.~Gupta.
\newblock Revisiting unreasonable effectiveness of data in deep learning era.
\newblock In {\em IEEE Int. Conf. on Computer Vision}, 2017.

\bibitem{sun2018developing}
Y.~Sun, X.~Zhang, Q.~Xin, and J.~Huang.
\newblock Developing a multi-filter convolutional neural network for semantic
  segmentation using high-resolution aerial imagery and lidar data.
\newblock {\em ISPRS Journal of Photogrammetry and Remote Sensing}, 2018.

\bibitem{zhao2017pyramid}
H.~Zhao, J.~Shi, X.~Qi, X.~Wang, and J.~Jia.
\newblock Pyramid scene parsing network.
\newblock In {\em IEEE Conf. on Computer Vision and Pattern Recognition
  (CVPR)}, pages 2881--2890, 2017.

\bibitem{zhou2018semievent}
Y.~Zhou, G.~Gallego, H.~Rebecq, L.~Kneip, H.~Li, and D.~Scaramuzza.
\newblock Semi-dense 3d reconstruction with a stereo event camera.
\newblock {\em ECCV}, 2018.

\bibitem{zhu2018realtimeevent}
A.~Z. Zhu, Y.~Chen, and K.~Daniilidis.
\newblock Realtime time synchronized event-based stereo.
\newblock {\em CVPR}, 2018.

\bibitem{zhu2018evflownet}
A.~Z. Zhu, L.~Yuan, K.~Chaney, and K.~Daniilidis.
\newblock Ev-flownet: Self-supervised optical flow estimation for event-based
  cameras.
\newblock {\em arXiv preprint arXiv:1802.06898}, 2018.

\bibitem{zhu2016beyond}
H.~Zhu, F.~Meng, J.~Cai, and S.~Lu.
\newblock Beyond pixels: A comprehensive survey from bottom-up to semantic
  image segmentation and cosegmentation.
\newblock {\em Journal of Visual Communication and Image Representation},
  34:12--27, 2016.

\end{thebibliography}
}
 
\end{document}